\documentclass[10pt,journal,compsoc]{IEEEtran}
\usepackage{amsmath}
\usepackage{enumitem}
\usepackage{graphicx}
\usepackage{cancel}
\usepackage{amssymb}
\usepackage{xcolor}
\usepackage{easyReview}
\usepackage{caption}
\usepackage{subcaption}

\renewcommand{\tilde}{\widetilde}

\usepackage[backend=biber]{biblatex}
\begin{document}

\title{A generative neural network model for random dot-product graphs}
\author{Vittorio Loprinzo
        and~Laurent Younes
\thanks{Loprinzo and Younes are at Johns Hopkins University, Department of Applied Mathematics \& Statistics.}
}

\markboth{April 2022}
{?}

\IEEEtitleabstractindextext{%
\begin{abstract}
We present GraphMoE, a novel neural network-based approach to learning generative models for random graphs. The neural network is trained to match the distribution of a class of random graphs by way of a moment estimator. The features used for training are graphlets, subgraph counts of small order. The neural network accepts random noise as input and outputs vector representations for nodes in the graph. Random graphs are then realized by applying a kernel to the representations. Graphs produced this way are demonstrated to be able to imitate data from chemistry, medicine, and social networks. The produced graphs are similar enough to the target data to be able to fool discriminator neural networks otherwise capable of separating classes of random graphs.
\end{abstract}

}

\maketitle
\IEEEdisplaynontitleabstractindextext

\section{Introduction}

In many applications of data science methods in medicine, computational vision, chemistry, and related fields, availability of data is often a limiting factor. In the case of graphical data, it is not always possible to obtain datasets containing large numbers of instances (graphs) from a specific class of interest. Brain connectivity data (produced, e.g., by functional MRI), for example, is limited by cost and subject availability when associated with diseases, and generally restricted to a few dozens of subjects. Chemical connectivity data \cite{AIDS,COX2MD,OHSU} is limited by the number of molecules and therefore naturally scarce. Many modern machine learning methods require more data than may be readily available, so it is desirable to create generative models for the distributions underlying such data. Such models can indeed be used for statistical inference, methods evaluation and validation. 

A large variety of random graph models are used in the literature, but most of those that are invariant by isomorphisms (which constitutes our focus in this paper) belong to the family of kernel graphs, in which a random variable $X_i$ is associated to node $i$ and an edge $(i,j)$ is created with probability $K(X_i, X_j)$ for some (possibly random) function $K$ that takes values in $[0,1]$ \cite{lovasz2012large,aldous1981representations, diaconis2007graph,bickel2011method}. The Erdos-Renyi model \cite{erdHos1960evolution} corresponds to the situation in which $K$ is constant (so that $X_i$ becomes irrelevant). In the stochastic block model \cite{holland1983stochastic}, the $X_i$'s are independent and take values in a finite set that determines ``communities''. Dot-product graphs \cite{young2007} are such that $X_i$ take values in the unit ball of an inner-product space and $K(X_i, X_j)$ is equal to the inner product between $X_i$ and $X_j$. Our model, presented in section \ref{sec:model} also belongs to this category, which presents the advantage that, once the kernel function $K$ is fixed, the modeling effort is reduced to the distribution of the node variables.  

Neural approaches for generative models have grown in popularity recently in part due to the successes of generative adversarial networks (GANs) \cite{goodfellow2014generative,arjovsky2017wasserstein} and variational auto-encoders (VAEs) \cite{kingma2013auto-encoding}. Such generative models are capable of producing new data (such as fake images of faces of nonexistent humans \cite{Bao_2017_ICCV}), cleaning up noisy data, and simulating unobserved phenomena in the sciences. However, only a few attempts have been made at using the GAN paradigm for graph data \cite{zhang2018end,de_cao2018molgan}. Other approaches include \cite{you2018graphrnn}, which is based on an autoregressive modeling of vectorized connectivity data or \cite{hallonquist2021graph}, which uses an exponential random graph model to implement maximum likelihood estimation for ``scene graphs'' associated with annotated images. 

As generative neural models have demonstrated their ability at modeling complex large dimensional random processes, they provide a natural resource for the modeling of the node process $X$ in kernel graphs and will provide the basis of our model. We will not use, however, the GAN or VAE learning paradigms, but rather implement a more classical parametric estimation approach, using moment estimators. These moments will be based on ``graphlets'', or subgraph counts \cite{bickel2011method,graphlets}, which are statistics that evaluate the number of induced subgraphs of small size in a given isomorphism class that are present in the random graph. These graphlets can be seen as an analog of polynomial moments for collections of binary random variables, and act as sufficient statistics in the characterization of the distribution of relabeling invariant distributions on graphs of infinite size \cite{lovasz2012large,diaconis2007graph}. Because these counts can be represented as sums of binary variables, they are also compatible with stochastic gradient descent strategies in the learning algorithm, as will be seen in section \ref{sec:learning}.

We will introduce, in the rest of the paper, GraphMoE, our neural-network-based approach to creating generative models for random graphs (section \ref{sec:model}) and associated training algorithm (section \ref{sec:learning}). We use small amounts of data to learn kernel-based models for random graph  distributions, which are then able to generate  unlimited amounts of artificial data. These generative models will then be evaluated in section \ref{sec:experiments} by comparing, in particular, real to artificial data. 

\section{Notation and Model}
\label{sec:model}

\subsection{Graphs and basic terminology}

A \textit{graph} is a pair $G = (V,E)$, where $V$ is a set of elements called \textit{vertices} and $E$ is a set of pairs of vertices. Graphs are used to represent pairwise interactions between objects, and are familiarly visualized as networks, with nodes representing the vertices and lines between the nodes representing edges. In a 
\textit{random graph}, the inclusion or exclusion of each possible edge in the graph is determined by the result of a binary random variable. If $V$ is ordered, say $V = (v_1, \ldots, v_n)$, the adjacency matrix of $G$, denoted $A_G$ is the binary matrix with entry $(i,j)$ equal to 1 if $(v_i,v_j)\in E$ and to 0 otherwise. If $V$ is a set of integers, we will always assume that it is listed in increasing order.

Two graphs $G = (V,E)$ and $G' = (V', E')$ are isomorphic (we will write $G\sim G'$) if there exists a bijection $f: V \to V'$ such that $\tilde f: (v_1, v_2) \mapsto (f(v_1), f(v_2))$ is also a bijection from $E$ to $E'$.

A graph $(V',E')$ is a sub-graph of $(V,E)$ if $V'\subset V$ and $E'\subset E$. It is an induced subgraph if $E' = \{(i,j)\in E: i, j\in V'\}$. We will denote by $G_{V'}$ the subgraph of $G$ induced by $V'$. 

Finally, if $V'$ is a set, we will make the abuse of notation $V' \subset G$ if $G = (V,E)$ and $V' \subset V$. 

In the following, we will use $[n]$ to denote the set $\{1, \ldots, n\}$ and $[n]_p$ the set of subsets of $[n]$ with $p$ elements. Letting $\mathbb K_n$ denote the complete graph with vertex set $[n]$, we will let $\mathcal K_n$ denote the family of all subgraphs of $\mathbb K_n$. Denoting by $\mathcal S_n$ the set of all $n!$ permutations of $[n]$, we associate to $s\in \mathcal S_n$ and $G = (V,E)\in \mathcal K_n$ the relabelled graph $s\cdot G$ with vertices $\{s(i): i\in V\}$ and edges $\{(s(i),s(j)): (i,j)\in E\}$, which provides a group action of $\mathcal S_n$ on $\mathcal K_n$.

We use the common convention of denoting random variables with capital letters and their realization using lower-case. However, in situations where  instances of  objects are usually denoted with capital letters (such as sets or graphs), we will use double-line fonts, such as $\mathbb A$ or $\mathbb G$. Collections or objects or variables will be denoted using bold-face letters.

\subsection{Graph model}
\subsubsection{Kernels}
In this work, we propose to develop models and training algorithms for kernel graphs.  To describe our model in full generality, we let $\mathcal Z = [0, +\infty)^d$ denote the $d$-dimensional positive Euclidean hyperquadrant and we consider a symmetric function $\phi: \mathcal Z \times \mathcal Z \to [0, 1]$ that we will call a kernel.  Examples of such kernels include
\[
\phi(z_1, z_2) = \frac{z_1^T z_2}{|z_1|\,|z_2|}
\]
where $|z|$ denotes the Euclidean norm of $z$,
\[
\phi(z_1, z_2) = \exp(-|z_1-z_2|^2)
\]
(Gaussian kernel) or 
\[
\phi(z_1, z_2) = \frac{(1+z_1^Tz_2)^c}{\sqrt{(1+z_1^Tz_1)^c(1+z_2^Tz_2)^c}}
\]
where $c$ is an integer (normalized polynomial kernel). Note that these examples correspond to reproducing kernels \cite{aronszajn1950theory}, and can therefore be  interpreted as inner products in possibly infinite-dimensional reproducing kernel Hilbert spaces, which will make our model a special instance of a {\em Random Dot Product Graph} (RDPG) \cite{young2007}. However, the reproducing property is not a required condition. For example, we obtain good numerical results when modeling sparse graphs and using
\[
\phi(z_1, z_2) = 1 - \frac{z_1^T z_2}{|z_1|\,|z_2|}\,.
\]

\subsubsection{A family of kernel graphs}
\label{sec:basic}
Fixing a kernel $\phi$, 
we will model graphs with random, but bounded, size, where the upper bound on the size is given by a fixed integer $n$. More precisely, our model will be supported by the set (denoted $\mathcal K_n$ above) of subgraphs of the complete graph $\mathbb K_n$.  
We will model below a collection of random variables $\boldsymbol Y = (\boldsymbol Q, \boldsymbol Z) = ((Q_1, \ldots, Q_n), (Z_1, \ldots, , Z_n))$ taking values in $[0,1]^n \times \mathcal Z^n$. Conditionally to a realization $\boldsymbol y = (\boldsymbol q , \boldsymbol z)$, we define random variables $\boldsymbol B, \boldsymbol R$, where
\begin{enumerate}
\item $\boldsymbol B = (B_1, \ldots, B_n) $ is a collection of independent Bernoulli variables with respective parameters $q_1, \ldots, q_n$.
    \item $\boldsymbol R = (R_{ij}, 1\leq i < j \leq n)$ is a collection of independent Bernoulli random variables with $P( R_{ij} = 1) = \phi(z_i, z_j)$.
    \item $\boldsymbol B, \boldsymbol R$ are mutually independent.
\end{enumerate}
We then define, for given realizations of $\boldsymbol B$ and $\boldsymbol R$,  the graph $G= \mathcal G(\boldsymbol b, \boldsymbol r)\in \mathcal K_n$ with vertex set $\{i: b_i = 1\}$ and edge set $\{(i,j): r_{ij} = 1, b_i=b_j=1\}$.

Our random graph model is then
\[
\mathbb G = S\cdot \mathcal G(\boldsymbol B, \boldsymbol R)
\]
where $S$ follows a uniform distribution on $\mathcal S_n$.
Equivalently, $\mathbb G$ is built by first forming a graph with vertex set $[n]$ and inserting an edge $(i,j)$ with probability $\phi(z_i, z_j)$, then selecting an induced subgraph by retaining each vertex with probability $q_i$ before randomly labelling the nodes. As required, this distribution is invariant by isomorphism, i.e., it gives the same probability to two isomorphic subgraphs of $\mathbb K_n$. 

Note that this graph construction differs from the one that is commonly used for RDPGs \cite{young2007,avanti}, which does not include a node selection mechanism, for which the graph size is generally large compared to $n$, and $Z_1, \ldots, Z_n$ are associated with ``communities'' that each node selects at random before forming edges. Our model and algorithms can however be easily modified to represent community graphs,  and we will return to this in section \ref{sec:community}.

\subsubsection{Latent variable model}
The latent variable $\boldsymbol Y$ is modeled as a function $\boldsymbol Y = \Phi(\Omega, \theta)$ where $\Omega$ is a random variable with known distribution (e.g., multivariate standard Gaussian) and $\theta$ is a vector of parameters. In our implementation $\Phi$ is specified by a neural net architecture of which $\Omega$ is the input layer and $\theta$ the vector of weights. However, a neural net implementation is by no means a necessity, and the further developments only require that $\Phi$ and its derivative in $\theta$  are reasonably easy to compute, for the learning algorithm to be feasible.

Our model is therefore fully specified by the kernel function, $\phi$, the maximal number of nodes, $n$, the ``architecture'' $\Phi$, and the parameter $\theta$. In the following, we assume that the first three are fixed and selected once for all, and focus on the estimation of $\theta$. 

Returning to the notation above, we will denote by $P_\theta$ (with expectation $E_\theta$) the distribution of the random graph before relabelling, i.e., that of $\mathcal G(\boldsymbol B, \boldsymbol R)$. The distribution of $S\cdot \mathcal G(\boldsymbol B, \boldsymbol R)$ is denoted $P^*_\theta$, with expectation $E^*_\theta$, so that, for $G\in \mathcal K_n$,
\[
P^*_\theta(G) = \frac 1{n!} \sum_{s\in \mathcal S_n} P_\theta(s\cdot G).
\]

\section{Learning algorithm}
\label{sec:learning}

\subsection{General setting}
We use a moment estimation approach, solving the equation 
$$E^*_\theta (H) = \bar{H},$$ 
where $H: \mathcal K_n \to \mathbb R^M $ is a vector of graph statistics, and $\bar{H}$ represents the empirical value of these statistics in the training data.

Since we want to learn a distribution that is invariant by graph isomorphism, $H$ should also have this property, i.e., $H(G) = H(G')$ if $G$ and $G'$ are isomorphic. Note that, if $H$ is relabelling invariant, one has
\begin{align*}
E^*_\theta(H) &= \sum_{G \in \mathcal K_n} H(G) P^*_\theta(G) \\
&= \frac1{n!} \sum_{s\in \mathcal S_n} \sum_{G \in \mathcal K_n} H(G) P_{\theta}(s\cdot G)\\
&= \frac1{n!} \sum_{s\in \mathcal S_n} \sum_{G \in \mathcal K_n} H(s^{-1}\cdot G) P_{\theta}(G)\\
&= \sum_{G \in \mathcal K_n} H(G) P_{\theta}(G)\\
&= E_{\theta}(H)\,.
\end{align*}
so that we can solve the simpler equation 
$$E_\theta (H) = \bar{H},$$
which will remove the relabelling step from our consideration.

\subsection{Graphlets}
Our function $H$ will be based on subgraph counts, or {\em graphlets}.
More precisely, let $F$ be a graph with $p$ vertices, and $G$ a subgraph of $\mathbb K_n$. Denote by $H_F(G)$ the probability that a set $A \in [n]_p$, chosen uniformly at random, is included in the vertex set of $G$ and induces a subgraph that is isomorphic to $F$.  For a graph $G$ with vertex set $V \subset [n]$,  we  have
\begin{align*}
H_F(G) &= \binom{n}{p}^{-1} \sum_{W \in [n]_p} \mathbf 1_{W \subset V} \mathbf 1_{G_W \sim F} \\
&= \binom{n}{p}^{-1} \sum_{W\subset V, |W| = p} \mathbf 1_{G_W \sim F},
\end{align*}
so that $H_F(G)$ is the number of induced subgraphs of $G$ that are isomorphic to $F$ normalized by $\binom{n}{p}$. This awkward probabilistic definition will make sense when we will discuss stochastic gradient descent.

We select a finite set $\mathcal F$ of graphs and define $H(G) = (H_F(G), F \in \mathcal F)$. Letting $\mathcal G_p$ denote the set of isomorphism classes of graphs with $p$ nodes, we use, in our experiments, $\mathcal F = \mathcal G_1 \cup \mathcal G_p$ for some integer $p$. ($\mathcal G_1$ is the trivial class of graphs with one node and the corresponding $H_F$ associated to $G$ its number of nodes divided by $n$.)  

\begin{figure}[h]
\centering
\includegraphics[scale=.45]{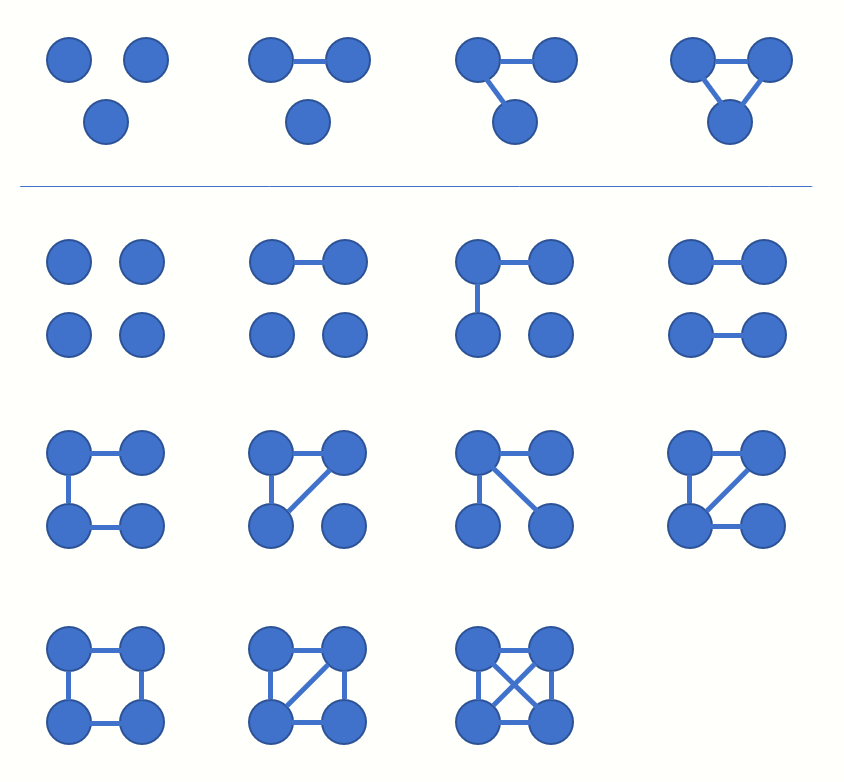}\caption{\label{fig:graphlets} On three nodes, there are four graph isomorphism classes, and on four nodes, there are eleven.}
\end{figure}

Equivalent classes of graphs are shown in Figure \ref{fig:graphlets} for $p=3$ and $4$. There are 34 isomorphism classes with 5 nodes, 156 with 6, more than 1,000 with 7 and more than 12,000 with 8 (note that, in this paper, we do not require $F$ to be connected in order to provide a graphlet).

Subgraph counts are polynomials in the binary edge variables of the random graph corrected for isomorphism invariance and therefore are generalizations of polynomial moments to random graphs. It is therefore not surprising that they constitute key elements in the asymptotic study of isomorphism-invariant graph distributions, as illustrated in \cite{lovasz2012large,diaconis2007graph}, for which they essentially provide sufficient statistics. Even though we are interested in finite, not necessarily large, graph models in this paper, using subgraph counts as the basis of our moment estimator remains natural, and they have proven
to be powerful also for the discrimination between different classes of graphs, in a wide range of applicative contexts \cite{zhang2011integrating,zhang2011integrating,jin2018network,janssen2012model}. 

\subsubsection{Objective function}
\label{sec:objective}
In the next section, we design a stochastic gradient descent (SGD) algorithm for the minimization of
\begin{equation}
\label{eq:obj}
U(\theta)  = (E_\theta(H) - \bar H)^TD(E_\theta(H) - \bar H)
\end{equation}
where $D$ is a fixed diagonal matrix of weights, with positive coefficients, the simplest choice being the identity matrix. We will write $\mathrm{diag}(D) = (D_F, F\in\mathcal F)$.

The first part of the training procedure therefor consists in estimating the target expectation $\bar H$ from training data (constituted, say, by graphs $G_1, \ldots, G_N$). It is in principle simply given by the empirical averages
\[
\bar H_F = \frac1N \sum_{k=1}^N H_F(G_k).
\]
However, the computation of $H_F(G)$ has polynomial complexity in the size of $G$, with a power equal to the size of $F$. The resulting cost can therefore be prohibitive for large graphs/graphlets,  and one must replace exact counts with approximations in the determination of $\bar H_F$. 

A notable body of work has been dedicated to the derivation of efficient algorithms for the approximation of subgraph counts (see, e.g., \cite{ribeiro2021survey} for a recent review), many of them optimized for the search of one specific graphlet rather than all graphlets of a given size. Since we here need all graphlets of size $p$, we have used the simple strategy of randomly sampling a sufficient number of subsets of nodes of cardinality $p$, identifying for each of them its isomorphism class to estimate frequencies. More explicitly, if $J$ subsets are sampled from $G_k$, and $J_F$ among them are found to be isomorphic with $F$, we estimate
\[
\hat H_F(G_k) = \frac{\binom{n_k}{p}}{\binom{n}{p}} \frac{J_F}{J}
\]
to approximate $H_F(G_k)$, where $n_k\leq n$ is the number of nodes in $G_k$.

\subsection{SGD formulation}
\label{sec:sgd}

We recall that our model defines a vector $\boldsymbol{Y}$ of latent variables generated as a function $\boldsymbol{Y} = \Phi(\theta, \Omega)$ where $\Omega$ is a random variable with known distribution, $\theta$ is a parameter, $\Phi$ is assumed to be differentiable in $\theta$ and $\boldsymbol{Y}$ takes the form
\[
\boldsymbol{Y} = ((Q_1,\ldots, Q_n), (Z_1,\ldots, Z_n)).
\]
Conditionally to $\boldsymbol Y$, we defined a node selection process $\boldsymbol B$ as a vector of independent Bernoulli variables with respective parameters $Q_1, \ldots, Q_n$ and an edge selection process  $\boldsymbol R$ as a triangular array of Bernoulli variables with parameters $\phi(Z_i, Z_j)$ yielding a graph $G = \mathcal G(\boldsymbol B, \boldsymbol R)$. In the following sequence of steps, we progressively express the objective function $U$ as the expectation, over an increasing number of random variables, of an increasingly simple function.

\begin{enumerate}[label=(\roman*),wide=0cm]
\item 
If we introduce two independent copies of $\Omega$, say, $\Omega, \tilde \Omega$, we can write
\begin{align*}
U(\theta) &= (E_\theta(H) - \bar H)^TD(E_\theta(H) - \bar H)\\
&= E(E_\theta(H - \bar H\mid\Omega))^T D E(E_\theta(H - \bar H\mid\Omega))\\
&= E(E_\theta(H - \bar H\mid\tilde \Omega)^TD(E_\theta(H - \bar H\mid\Omega))
\end{align*}
where the outer expectation is with respect to the distribution of $\Omega, \tilde \Omega$. 
\item Moreover, we have
\[
H_F(G) = P_{n,p}(\mathbb W \subset G \text{ and } G_{\mathbb W}\sim F)
\]
where $P_{n,p}$ denotes the uniform distribution over random subsets, $\mathbb W$, of $[n]$ with cardinality $p$,
so that, introducing the set $\tilde{\mathcal J}_{W,F}$ of graphs $G\in \mathcal K_n$ such that $W\subset G$ and $G\sim F$, we have
\[
E_\theta(H\mid \Omega = \omega) = E_{n,p}(P_\theta(\tilde{\mathcal J}_{\mathbb W, F} \mid \Omega = \omega)).
\]
As a consequence, we can write
\[
U(\theta) = E((\zeta(\theta, \tilde \Omega, \tilde {\mathbb W}) - \bar H)^TD(\zeta(\theta, \Omega, \mathbb W) - \bar H))
\]
where $E$ now represents an expectation with respect to $\Omega, \tilde\Omega, \mathbb W, \tilde {\mathbb W}$ and $\zeta(\theta, \omega, W)$ is the vector formed by
\[
\zeta_F(\theta, \omega, W) = P_\theta(\tilde{\mathcal J}_{W,F} \mid \Omega = \omega), f\in \mathcal F.
\]
\item 
Let $F$ be a graph with vertex set $[p]$.
Let $\mathcal A_F$ denote the set of possible adjacency matrices of graphs isomorphic to $F$, i.e., the set of all distinct matrices $(A_F(s(i),s(j)))$ where $A_F$ is the adjacency matrix of $F$ and $s\in \mathcal S_n$. 

Then, letting, for an adjacency matrix $A$, $\mathcal J_{W,A}$ denote the set of graphs $G\in \mathcal K_n$ such that $W\subset G$ and $A_{G_F} = A$, we have
\[
P_\theta(\tilde{\mathcal J}_{W,F}\mid \Omega = \omega) = \frac{1}{|\mathcal A_F|} \sum_{A\in\mathcal A_F} \eta(\theta, \omega, W, \tilde A)
\]
where
\[
\eta(\theta, \omega, W, A) = |[A]| P_\theta(\mathcal J_{W,A}\mid\Omega = \omega),
\]
where $[A]$ denotes the isomorphism class of $A$, and we have $|[A]| = |\mathcal A_F|$ if $A \in \mathcal A_F$.
\item We have
\begin{align*}
& (\zeta(\theta, \tilde \omega, \tilde { W}) - \bar H)^T D (\zeta(\theta, \omega, W) - \bar H)
= \\
& \,\,\,= \sum_{F\in \mathcal F} D_F (\zeta_F(\theta, \tilde \omega, \tilde {W}) - \bar H_F)(\zeta_F(\theta, \omega, W) - \bar H_F)\\
& \,\,\,= \sum_{F\in \mathcal F} D_F |\mathcal A_F|^{-2}\\
& \qquad \sum_{A,\tilde A\in \mathcal A_F} (\eta(\theta, \tilde \omega, \tilde {W}, \tilde A) - \bar H_F)(\eta(\theta, \omega, W, A) - \bar H_F)\\
& \,\,\,= \mathrm{tr}(D) E\left((\eta(\theta, \tilde \omega, \tilde {W}, \tilde {\mathbb A}) - \bar H_{\tilde{\mathbb A}})(\eta(\theta, \omega, W, \mathbb A) - \bar H_{\mathbb A})\right)
\end{align*}
where we have introduced the random variables $(\mathbb A,  \tilde {\mathbb A})$ whose joint distribution is as follows: first choose a graphlet class $\mathbb F$ over $\mathcal F$ with probability proportional to $D_F$ and, conditionally to $\mathbb F = F$, take $\mathbb A$ and $\tilde {\mathbb A}$ independent and both uniformly distributed over $\mathcal A_F$. Here, we made the abuse of notation $\bar H_A := \bar H_F$ for any $F$ with adjacency matrix $A$.

\item This leads to our SGD implementation, that computes a sequence of parameters $(\theta_t, t\geq 1)$ using
\begin{multline}
\label{eq:sgd}
\theta_{t+1} = \theta_t -
\gamma_t \sum_{i=1}^L \sum_{j=1}^M \partial_\theta \Big((\eta(\theta, \tilde \omega^{(j)}, \tilde W^{(i)}, \tilde A) - \bar H)^T\\
(\eta(\theta, \omega^{(j)}, W^{(i)}, A) - \bar H)\Big)
\end{multline}
where $\gamma_t$ is the learning rate, $\omega^{(1)}, \ldots, \omega^{(M)}$, $\tilde \omega^{(1)}, \ldots, \tilde \omega^{(M)}$, $W^{(1)}, \dots, W^{(L)}$, $\tilde W^{(1)}, \ldots, \tilde W^{(L)}$ are independent samples of $\Omega$ and $\mathbb W$ and $A, \tilde A$ independent samples of $\mathbb A$. 

To complete the presentation of the SGD algorithm, we need make explicit the computation of $\eta(\theta, \omega, W, A)$ and its derivative in $\theta$, as a function of the derivatives of $\Phi$ in $\theta$ that are supposed to be computable either explicitly or using known algorithms (such as back-propagation).
\end{enumerate}

\subsection{Conditional expectations of graphlets}
\label{sec:cond.exp}

Let $W\in [n]_p$. We let $k_i\in [p]$ denote the rank in $W$ of one of its elements, $i$.  
Then, a graph $G = \mathcal G(\boldsymbol b, \boldsymbol r)$ is such that $W\subset G$ and $G_W \sim F$ if and only if $b_{j}=1$ for all $j\in W$ and there exists $A \in \mathcal A_F$ such that $r(i,j) = A(k_i,k_j)$ for $i,j\in W$, $i<j$.  This shows that
\begin{align*}
&\eta(\theta, \omega, W, A) = |[A]| P_{\theta} (\mathcal J_{W,A}|\Omega=\omega) \\
&=|[A]| \prod_{i\in W} q_{i} \prod_{\stackrel{i,j\in W}{i<j}} \phi(z_i, z_j)^{A(k_i,k_j)} (1- \phi(z_{i}, z_{l}))^{1-A(k_i,k_j)} .
\end{align*}
Denote the right-hand side by $\psi_A(\boldsymbol q, \boldsymbol z)$. Then 
\[
\partial_\theta \eta(\theta, \omega, W) = \sum_{i\in W} (\partial_{q_i} \psi_A \partial_\theta q_i + \partial_{z_i} \psi_A \partial_\theta z_i)
\]
where, we recall, $(\boldsymbol q, \boldsymbol z) = \Phi(\theta, \omega)$ whose derivative in $\theta$ are assumed to be known. So, only the derivatives of $\psi_A$ in $\boldsymbol q$ and $\boldsymbol z$ need to be made explicit, and since $\psi_A$ is polynomial in these variables, the computation is elementary, with
\[
\partial_{q_i} \psi_A(\boldsymbol q, \boldsymbol z) = \frac{1}{q_i} \psi_A(\boldsymbol q, \boldsymbol z)
\]
and, letting
\begin{multline*}
\Lambda_{A, W} (\boldsymbol q, \boldsymbol z) = \\
    |[A]| \prod_{i\in W} q_{i} \prod_{\stackrel{i,j\in W}{i<j}} \phi(z_{i}, z_{j})^{A(k_i,k_j)} (1- \phi(z_{i}, z_{j}))^{1-A(k_i,k_j)},
\end{multline*}
we have
\begin{multline*}
\partial_{z_{i}} \psi_A(\boldsymbol q, \boldsymbol z) =\\
\sum_{\stackrel{j\in W}{j\neq i}} \left( \frac{\partial_{z_{i}} \phi(z_i,z_j)}{\phi(z_{i},z_{j})} A(k_i,k_j) - \frac{\partial_{z_{i}} \phi(z_i,z_j)}{1-\phi(z_{i},z_j)}(1-A(k_i,k_j))\right) \\
\times \Lambda_{A, W} (\boldsymbol q, \boldsymbol z).
\end{multline*}

\subsection{Extensions}
\label{sec:extensions}
The formulation of the SGD algorithm, leading to equation \eqref{eq:sgd},  relies on the fact that subgraph counts were used for the moment estimators, but not on the specific stochastic model used to generate the graphs.  The approach that it suggests can  be applied to graph models that differ from the one we have introduced and therefore defines a general learning strategy for random graphs.
The feasibility of the approach however depends on whether the functions that we have denoted by $\eta(\theta, \omega, W, A)$ in section \ref{sec:sgd} and their derivatives are easy to compute, as described in section \ref{sec:cond.exp} for our model. This would be the case for any model that would define a latent variable $\mathbf Y = \Phi(\omega, \theta)$, like ours, conditionally to which vertex selection and edge insertion are modeled as independent variables. We now discuss a few examples.

\subsubsection{Adjacency matrix model}
Instead of using a kernel graph, one may choose to directly model the adjacency matrix of the random graph. To simplify the discussion, we restrict to situations when generated graphs have a fixed size (therefore not including a vertex selection step). In this model, the latent variable  $\boldsymbol Y$ is a symmetric matrix  with entries in the unit interval, and, conditionally to $\boldsymbol Y= \boldsymbol y$, an edge  $(i,j)$ is included in the graph with probability $y_{ij}$ (before random relabelling). In that case, 
\begin{multline*}
\eta(\theta, \omega, W, A) =\\
|[A]|\prod_{i\in W} q_{i} \prod_{\stackrel{i,j\in W}{i<j}} \phi(z_i, z_j)^{A(k_i,k_j)} (1- \phi(z_{i}, z_{l}))^{1-A(k_i,k_j)},
\end{multline*}
which is still a polynomial in $\phi$. We have opted for using kernel graphs models rather than full adjacency matrices in our main model because the former are more parsimonious, in terms, in particular, of the dimension of the generative network, while still offering a wide modeling range. Moreover, they implicitly provide a linear embedding of the generated graph, which has clear advantages for data analysis.

\subsubsection{Community graphs}
\label{sec:community}
Community models of random graphs may also be generated using our approach. In this case, rather than pairing representations to nodes, they are paired to communities, of which only a small fixed number, say, $t$ are learned. Assuming that $t$ is known, the corresponding latent variable is a collection of $t$ vectors, $\boldsymbol Y = \boldsymbol Z = (Z_1, \ldots, Z_t)$. 
We also model an intermediary random variable $\boldsymbol{C}$ that assigns nodes to communities, such that (assuming $n$ nodes) $\boldsymbol C = (C_1, \ldots, C_n)\in [t]^n$ with $C_1, \ldots, C_n$ independent and identically distributed. Letting $s_c = P(C_i = c)$, 
we now have
\begin{multline*}
\eta(\theta, \omega, W, A) = |[A]| \times \\
\sum_{c \in [t]^{|W|}} \prod_i s_{c_i} \prod_{\stackrel{i,j\in W}{i<j}} \phi(z_{c_i}, z_{c_j})^{A(k_i,k_j)} (1- \phi(z_{c_i}, z_{c_j}))^{1-A(k_i,k_j)}
\end{multline*}
Note that the $s_c$'s are model parameters and therefore also need to be learned. Importantly, $\eta$ is polynomial in $s$ and $\phi$, so that the previous SGD approach can be used, with the following simplification.

We recall that, in the previous SGD formulation, a set of nodes $\mathcal W$ was selected uniformly at random over all nodes in $G$, where the size of $W$ is equal to the size of the graphlet being considered. In the community model described here, selection of $\mathcal W$ is equivalent to selecting a set of the representations produced by the model with replacement, since nodes now share representations. 
Adjacency matrices $\mathbb A$ and $\tilde {\mathbb A}$ were also chosen uniformly at random over $\mathcal A_F$. The set $\mathcal A_F$ is a permutation class of matrices, i.e., for each pair of matrices $A$, $\tilde{A} \in \mathcal A_F$, there exists a permutation matrix $V$ such that $\tilde{A} = VAV^T$, and $\mathcal A_F$ is also closed under such permutations.

There is now redundancy between the above sum and the choices of $A$ and $\tilde{A}$. Since the sum is over all possible choices of representations for the nodes in $W$, all permutations of a choice of  representations are included in the sum as separate terms. Since, by sampling $\mathcal A$ in the previous SGD algorithm, we were only selecting a permutation, this step is no longer necessary with the community model. Computing the sum comes with computational cost, but removes one of the stochastic components from the algorithm. 

The  computation of the derivatives required for stochastic gradient descent is largely unchanged, and the derivatives for the new $s$ parameters are trivial.

\subsubsection{Partial graphlets}
\label{sec:partial}
When dealing with graphs that have heavy-tailed degree distributions, large edge combinations (i.e., large graphlets) may be needed to accurately model the data. However, due to the large number of graphlets of order larger than six, and of their probable scarcity in a finite dataset, it is necessary to select them and group them before computing a moment estimator. 

We here introduce \textit{partial graphlets}, that we define  to be a graphlet where a subset of the edges are ignored. More formally, we define a partial graphlet of order $p$ by a $p \times p$ matrix $M$ whose entries are 1, 0, or $-1$. We then say that a subgraph $G$ of size $p$ with adjacency matrix $A$ is consistent with a partial graphlet if, for some permutation matrix $V$,
\[ (VAV)_{ij} = M_{ij} \,\,\, \text{  or  }\,\,\, M_{ij} = -1, \,\,\, \forall i,j \in [p]. \]
This operation groups together graphlets of some maximum size that contain a specific pattern, avoiding the issue of seeing their   probabilities shrinking too small. 

In particular, we make use of the ``star'' partial graphlet:
\[
M_{ij} = \begin{cases}
0 & i = j \\
1 & i = 1, \,\, j \neq i\\
-1 & \text{otherwise}
\end{cases}
\]
At values of $p$ larger than previously employed in this work, this partial graphlet is closely related to the probability of a node having high degree. Whether a chosen subgraph is consistent with this partial graphlet is also simple to check through a  row sum of the subgraph's adjacency matrix.

\section{Examples of kernels}
\label{sec:kernels}
The graph kernel, $\phi$, is an essential component of our model. Its selection is an important step when designing a model, and requires  some basic understanding of the dataset being modeled. Useful information includes the average number of vertices, the distribution of vertex degrees, and any community structure information that may be available. Here, we list kernels that were used in experiments, including a discussion of strengths and weaknesses of each. Each kernel $\phi(z_1, z_2)$ acts as $\phi: \mathbb{R}^d \times \mathbb{R}^d \to [0, 1]$.

\subsection{Dot product kernel.} The dot product kernel is 
\[ \phi(z_1,z_2) = \frac{|z_1^T z_2|}{|z_1| | z_2|}.\]
This is among the most basic kernels, and is also that most commonly used when RDPGs are discussed. For small values of $d$, it has the advantage of training very quickly because of its simple functional form. However, if the graphs are sparse, the node representations $z_i \in \mathbb{R}^d$ must be nearly mutually perpendicular, requiring $d$ to grow with the number of nodes in the produced graphs. For this reason, the dot product kernel is effective for small graphs (or small numbers of communities in the case of community graphs), but becomes inefficient as the graph size grows.

\subsection{Complement dot product kernel.} This kernel is simply the complementary probability of that produced by the dot product kernel, ie. 
\[ \phi(z_1,z_2) = 1- \frac{|z_1^T z_2|}{|z_1| | z_2|}.\]
If the graph contains large groups of nodes with very small probabilities of edges being formed between them, this kernel is highly effective. It was designed specifically as a solution for generating bipartite graphs.

\subsection{Radial basis function kernel.} The RBF or Gaussian kernel is 
\[ \phi(z_1,z_2) = \exp (-|z_1-z_2|^2). \]
When this kernel is used, the probability of an edge between two nodes is dependent only on the proximity of their representations. Because of this, many more probabilities can be accurately modeled without the need for large $d$; for this reason, it vastly outperformed the dot product kernel in both accuracy and speed of training on all datasets with large numbers of nodes. 

\subsection{RBF kernel with scale factor.} This is a variation on the previous kernel:
\[ \phi(z_1,z_2) = \sqrt{(1+|z_1|^2)(1+|z_2|^2)} \exp (-|z_1-z_2|^2). \]
With the unscaled RBF kernel, if a set of nodes all mutually have edge probability $p < 1$, all of their representations must exist at the same mutual distance from each other, which is difficult to achieve in low dimension $d$. By adding the scale factor, the set of nodes can all share a representation $z^*$, as long as that representation's distance to the origin creates edge probability $\phi(z^*, z^*) = p$. As such, this kernel is especially effective for graph distributions with community structure (for which, however, the variant described in section \ref{sec:community} may be preferred). 

Because this kernel is not invariant to the scale of the inputs, we found the most success when adding a penalty to the objective function controlling the size of the network's weights, i.e., 
\[
U(\theta) = |E_\theta(H) - \bar{H}|^2 + \lambda g(\theta),
\]
with 
\[ g(\theta) = \begin{cases}
0 & |\theta| < \kappa \\
|\theta-\kappa|^2 & |\theta| \geq \kappa 
\end{cases}
\]
for some constants $\lambda$ and $\kappa$.
This ensures that the network outputs do not grow too large, avoiding, in particular, the region where the derivative of the kernel is close to 0.

\subsection{Polynomial kernel.} We include this example for completeness, as it was not used in our experiments. The normalized version of the polynomial kernel is 
\[\phi(z_1, z_2) = \frac{(1+z_1^Tz_2)^c}{\sqrt{(1+z_1^Tz_1)^c(1+z_2^Tz_2)^c}}
\]
for some integer $c$. This kernel again depends on the inner product of the inputs, similarly to the dot product kernels. Its nonlinearity in the inner product avoids the issue of dimensionality described above to some extent, but the kernel still suffers when large numbers of nodes all share a mutual edge probability. While it offers great flexibility, we found less practical use for this kernel than the dot product and RBF variations.

\section{Implementation}
\label{sec:implementation}

In addition to the kernel discussed in the previous section, the training algorithm needs to be provided with the maximum number of nodes, $n$, maximum size of graphlets used for learning, the geometry of the neural network used to train the function $\Phi$, the distribution of its input $\Omega$ and the weight matrix $D$ used in the definition of the objective function. Additional inputs can also specify some parameters of the SGD procedure, such as the number of examples used in minibatches.

In our implementation, a library of graphlets of size up to $p=8$ was pre-computed and an array of size $p(p-1)/2$ was built and stored for fast retrieval of the graphlet class  associated with a $p\times p$ binary connectivity matrix. 
We used a feed-forward fully connected neural network with two hidden layers for the function $\Phi$. Its input layer, forming the variable $\Omega$, was generated as independent standard Gaussian variables. The activation function was Leaky ReLU, i.e.,
\[
\tau(x) = \begin{cases}
x & x\geq 0 \\
\epsilon x & x <0
\end{cases}
\]
with $\epsilon = 0.01$.
The parameter $\theta$ is associated with the network's weights. The implementation was programmed using PyTorch, allowing for a straightforward computation of the derivatives of $\Phi$ with respect to
$\theta$.

The maximum number of nodes, $n$, was chosen  to be the maximum number of nodes in any graph in the dataset in use. 

While training, a step-down step size strategy was employed. Fix a number of phases, $\kappa$, target values of cost $u_1, \ldots, u_\kappa$ and step sizes $\gamma_{0} > \cdots > \gamma_{\kappa-1}$. We run the algorithm with step size $\gamma_0$ until the first time $t_1$ such that $U_{t_1}(\theta) < u_1$. We then use $\gamma_1$ until the first time $t_2>t_1$ such that $U_{t_2}(\theta) < u_2$ and so on until time $t_{\kappa}$ is reached and the procedure is stopped. (We used $\kappa=3$ in our experiments.) For this purpose, $U(\theta)$ was computed at regular intervals of training time.

\section{Datasets}
The algorithm was tested on a variety of datasets. It was first applied to a number of synthetic datasets, generated from popular generative models for random graphs. It was then tested on a series of real datasets with different characteristics, several of which were accessed from a repository at \cite{benchmark}. Details of each dataset are given below and in Table 1.

\subsection{Empty graph}
As a first step and sanity check, we attempt to imitate the empty graph distribution, i.e., graphs that always have zero edges. With the dot product kernel, this requires all representations to be mutually perpendicular. With the RBF kernel, the representations just must be as isolated as possible in Euclidean space.

\subsection{Stochastic Block Models}
\label{sec:sbm}
We attempt to learn the distribution of a class of graphs generated from a particular stochastic block model (SBM). In an SBM, each node is randomly assigned to one of several ``blocks,'' and it is the block memberships of a pair of nodes that determine the probability of an edge existing between them. 
Two SBMs were chosen. The first is the two-block model described in \cite{you2018graphrnn} (there called ``Community'') for comparison purposes. The second is a four-block model with higher within-block edge probabilities so that clustering of the representations can be cleanly visualized.
For the two-block model, the membership probabilities for each block are 
$$\pi = [0.5, 0.5].$$
The probability of an edge existing between a node from block $i$ and a node from block $j$ is given by the $i$\textsuperscript{th}, $j$\textsuperscript{th} entry of the matrix 
$$B = \begin{pmatrix}
0.3 & 0.05 \\
0.05 & 0.3
\end{pmatrix}.$$
For the four-block model, the membership probabilities for each block are 
$$\pi = [0.25, 0.25, 0.25, 0.25]$$
and the community probability matrix is 
$$B = \begin{pmatrix}
0.75 & 0.1 & 0.1 & 0.1 \\
0.1 & 0.75 & 0.1 & 0.1 \\
0.1 & 0.1 & 0.75 & 0.1 \\
0.1 & 0.1 & 0.1 & 0.75
\end{pmatrix}.$$
It is simple to show that if $B$ is positive semidefinite, the SBM is a submodel of the RDPG \cite{avanti}, so the method described above should be able to imitate the distribution produced by these SBMs.

\subsection{AIDS}
This is a chemical dataset, sourced from \cite{AIDS}. It consists of graphs from two classes: Compounds that are active against the HIV virus, and compounds that are not. Each graph represents a single chemical compound, where nodes represent atoms and edges represent covalent bounds between the atoms. We attempt to learn the ``Active'' class of graphs, and use the other class for comparison.

\subsection{COX2\_MD}
This is a publicly available chemical dataset, converted to a graph format by \cite{COX2MD}. Each graph represents a cyclooxygenase-2 inhibitor. Nodes represent atoms and edges exist between nodes if they are sufficiently close together in physical space.

\subsection{OHSU}
This dataset was constructed by \cite{OHSU} out of BrainNet Functional Brain Network Analysis Data. Here, nodes parcellate regions of the brain, and edges represent correlations between the regions. The two classes in this set represent typical brains and brains with Hyperactive-Impulsive classification.

\subsection{Brain}
Provided by \cite{carey} and first used in \cite{Lawrence2021, desikan}, this dataset consists of diffusion MRI data. Each graph is a component of a connectome, where nodes represent regions of the brain and edges exist between nodes when there are fiber streamlines connecting the regions.

\subsection{Protein}
Used in \cite{you2018graphrnn}. Here, each graph represents a protein, comprised of amino acids. Each node represents an amino acid, and two nodes are connected if their amino acids are physically less than six angstroms apart in the protein structure. 

\subsection{IMDb}
Used in \cite{IMDb}. The dataset consists of ego graphs created using the IMDb network. In the IMDb network, nodes represent actors and an edge exists between two actors if they appeared in the same film. In an ego network, there is one central node (the ``ego'') around which the rest of the network is constructed. In this case, a particular actor was chosen as the ego for each graph in the dataset, and any node connected to the ego was included in the graph. All edges between selected nodes were included as well.

\begin{table*}[h]
\centering
\begin{tabular}{l|lrrrrrr}
Dataset & Source & Num. Graphs & Avg. Nodes & Avg. Edges & \multicolumn{1}{l}{Tot. Diff.} & Max Diff. & \multicolumn{1}{l}{Kernel} \\ \hline
Empty graph & N/A & N/A & 10 & 0 & 0.000 & 0.000 & RBF3 \\
4-block SBM & N/A & N/A & 16 & 45 & 0.024 & 4.3e-3 & DP4 \\
2-block SBM & N/A & N/A & 80 & 548 & 0.067 & 8.8e-3 & RBF4 \\
AIDS & \cite{AIDS} \cite{AIDS2} & 2000 & 15.69 & 16.20 & 0.015 & 1.8e-3 & DP5 \\
COX2\_MD & \cite{COX2MD} & 303 & 26.28 & 335.12 & 0.045 & 5.2e-3 & DP5 \\
OHSU & \cite{OHSU} & 79 & 82.01 & 199.66 & 0.052 & 8.8e-3 & RBF5 \\
Brain & \cite{carey} & 114 & 70 & 1037 & 0.101 & 7.5e-3 & RBF5\\
Protein & \cite{you2018graphrnn} & 1113 & 39.05 & 72.82 & 0.071 & 8.2e-3 & RBF5 \\  
IMDb & \cite{IMDb} & 1000 & 19.77 & 96.53 & 0.122 & 9.3e-3 & RBF5 \\
\end{tabular}
\label{tab:my-table}
\caption{Description of the datasets used to test the generator. The first three are artificial, created from a known distribution. ''Total Difference'' is the total absolute difference between the target graphlets and the generator graphlets after training, with an optimal value of 0. ''Max Difference'' is the largest difference for a single graphlet between the target and generator. The best-performing kernel for each is listed: DP for dot product, RBF for radial basis function. The number following is the order of graphlets used to produce results.}
\end{table*}

\section{Results}
\label{sec:experiments}
\subsection{Basic Model}
We start with some results and validation of our basic model presented in section \ref{sec:basic}. For this first set of results, the matrix $D$ in \eqref{eq:obj} is the identity matrix.

\subsubsection{Training performance}
Table \ref{tab:results} provides some details on the outcome of the training algorithm in terms of correctly learning the graphlet expectations.

The results of the base algorithm's attempts to learn various random graph distributions were favorable. On the simplest examples, the algorithm is able to complete its task quickly and perfectly. When asked to learn the empty graph distribution, the algorithm is able to produce the desired results perfectly within a few hundred iterations. 

\begin{figure}[h]
\centering
\includegraphics[width=0.49\textwidth]{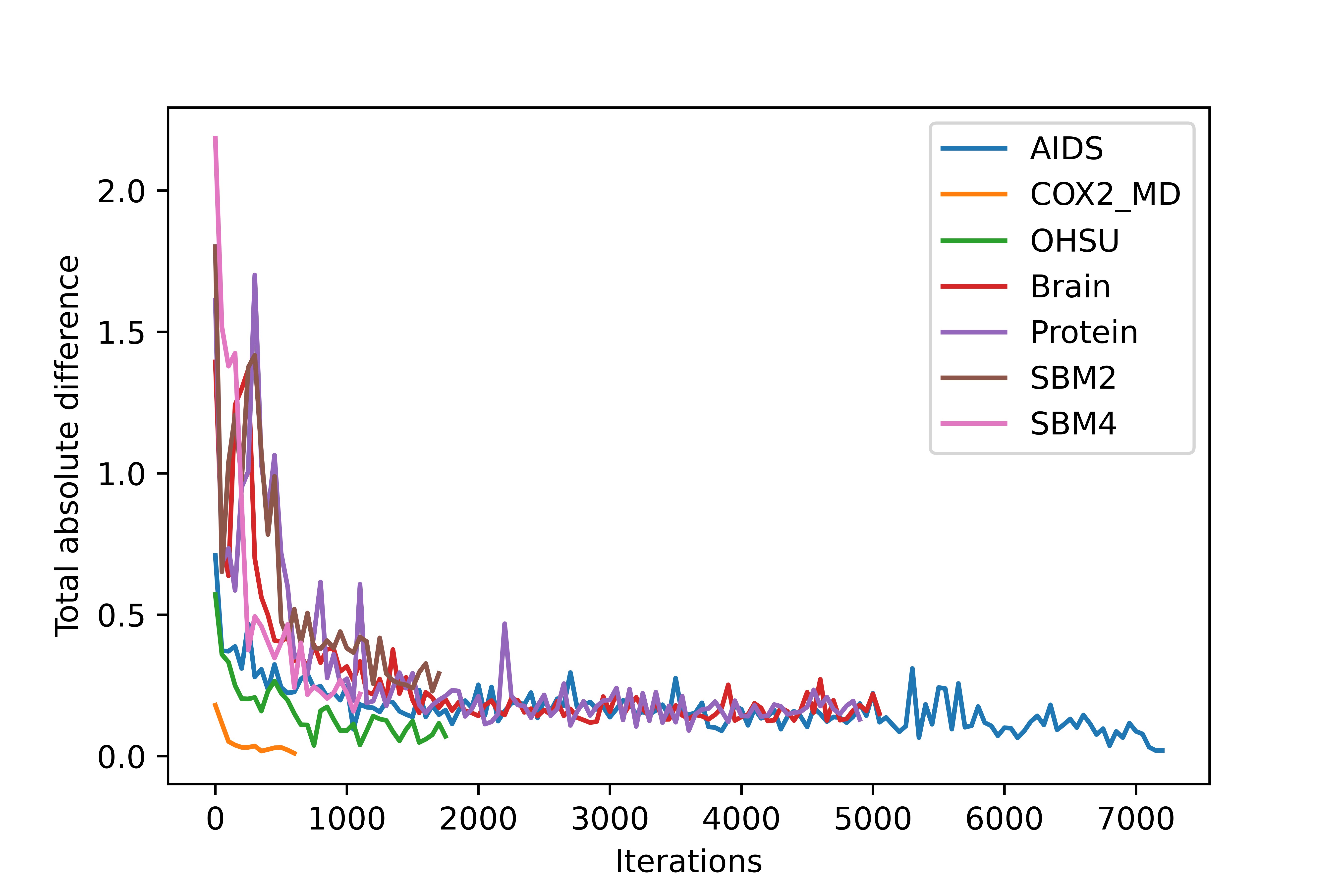}
\caption{\label{fig:3}
Training error for the GraphMoE generator for each dataset.}
\end{figure}

When attempting a more complicated example, such as  the four-block stochastic block model described in section \ref{sec:sbm}, success can be found with the right combination of hyperparameters, such as the number of nodes in the produced graphs, the dimension of their representations, the minibatch size, and the number of neurons in the neural network. In the case of the four-block SBM described above, ten-dimensional representations were used to create graphs on 80 nodes. Each hidden layer of the NN contained ten neurons. Within a few thousand iterations, the algorithm was able to match the graphlet distribution from the SBM to within 0.01\%.

The embeddings that resulted from the application of the algorithm in the case of the four-block SBM were separated into four clusters, with each cluster corresponding to a block in the SBM. Nodes whose embeddings are close together are more likely to have en edge between them, and thus should be members of the same block in the SBM. A two-dimensional projection of the embeddings produced using principal component analysis can be seen in Figure \ref{fig:3}.

\begin{figure}[h]
\centering
\includegraphics[width=0.5\textwidth]{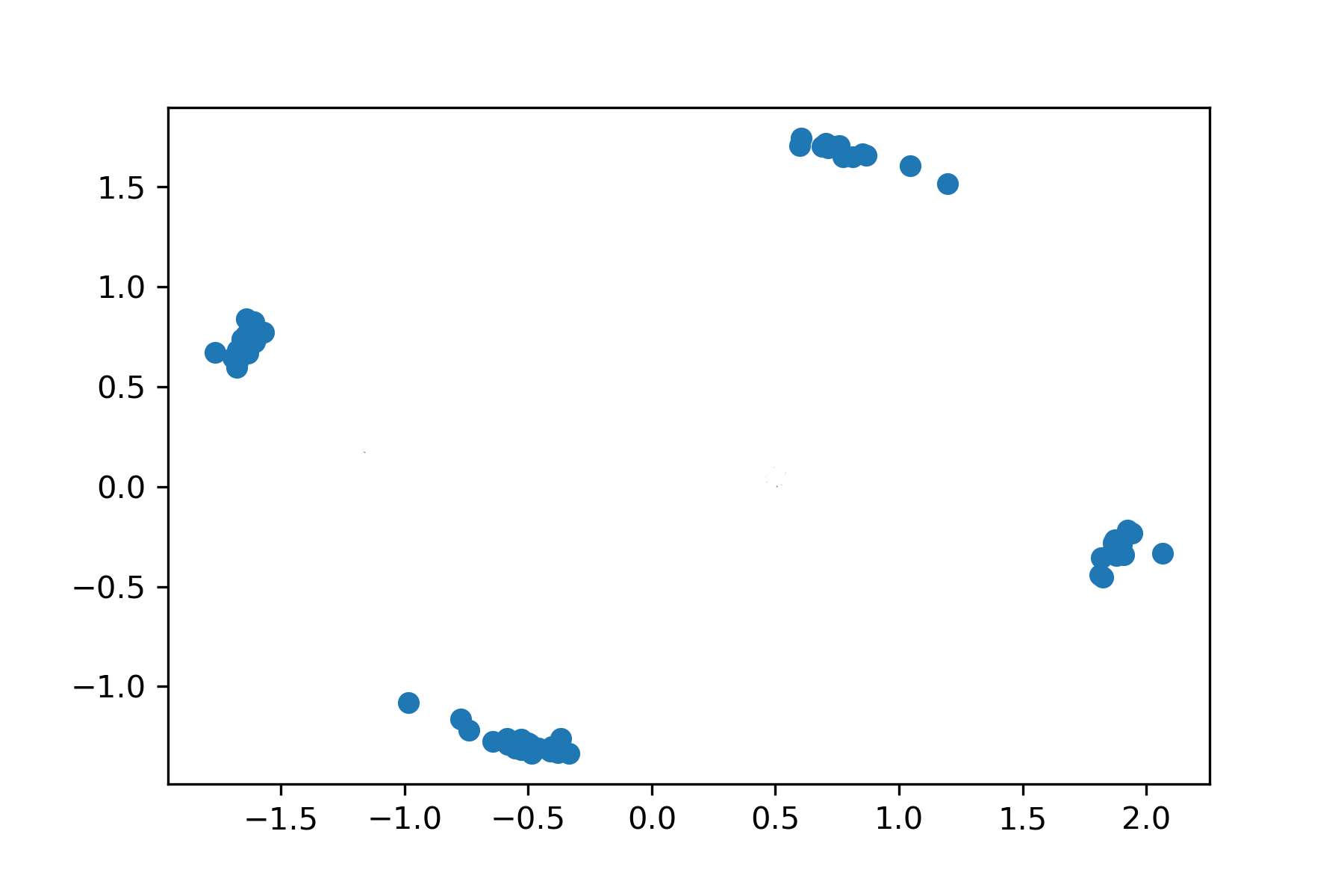}
\caption{\label{fig:4}
PCA visualization of node representations for four-block SBM trained using an RBF kernel.}
\end{figure}

\begin{figure}[h]
\centering
\includegraphics[width=0.35\textwidth]{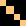}
\caption{\label{fig:heatmap}
The model was trained on the 4-block SBM. This matrix shows the frequency with which each node fell into the same community as each other node. Four clear communities are established, with zero mixing between them.}
\end{figure}

After running these test cases, the algorithm was used to learn the distributions of the several real datasets listed above. 
In these cases, the algorithm quickly converges to a solution, closely matching the desired graphlet proportions from the target distribution. The metric for error in these cases is the total absolute difference between the target graphlet proportions and the graphlet proportions of graphs produced by the network midway through training. The algorithm could usually reduce the absolute difference to less than 5\% of the overall graphlet proportions.

\textbf{Remark.} An examination of the typical output of the trained neural network reveals a shortcoming of this model. When training for a dataset with community structure, such as the SBM dataset, the embeddings cluster together, which is desirable. However, embeddings tend to become ``stuck'' in their clusters, and do not typically move between clusters when different inputs are passed through the net. As a result, in the case where the number of nodes in the produced graphs are fixed (ie. $q_i = 1\,\,\, \forall i$), each community will always contain the same number of nodes. This does not properly capture the behavior of the SBM or other community models, where nodes are randomly assigned to communities each time a graph is realized. Figure \ref{fig:heatmap} was produced by a net trained on the SBM described above; it is a heatmap visualization of a matrix whose $i,j^{th}$ entry is the frequency that the $i^{th}$ node fell in the same community as the $j^{th}$ node. This shortcoming limits the usefulness of the model when applied to certain types of datasets; however, using our community model from \ref{sec:community} properly addresses the issue (see section \ref{section:community}).

\subsubsection{Validation}

In order to further test the validity of the algorithm, we evaluated the performance of neural net classifiers in separating true from generated data in our datasets, where poor performance in classification is an indicator of the quality of the generative model.

Two NN classifiers were constructed. The first is a standard feed-forward neural network with two fully-connected hidden layers. The features used by this network were the graphlets of order one larger than those used to train the generators. These features were obtained for graphs in the datasets using the sampling technique described in section \ref{sec:objective}. The resulting classifier is powerful enough to  distinguish with near perfect accuracy between the two real classes from the AIDS, COX2\_MD, or OHSU datasets.

We used Graph Neural Nets (GNNs) as second classifier, because they share less with the generator in terms of features and architecture. GNNs, which are now fairly ubiquitous for graph classification tasks \cite{gnn1,gnn2}, make use of convolutional operations specially suited for graph data. A simple GNN was employed here, with two hidden graph-convolutional layers, and node degrees used as the input features. The GNN was less effective than the graphlet-based classifiers at distinguishing between classes of the real datasets, but still was able to discriminate effectively enough for testing purposes.

\begin{figure}[h]
\centering
\includegraphics[width=0.49\textwidth]{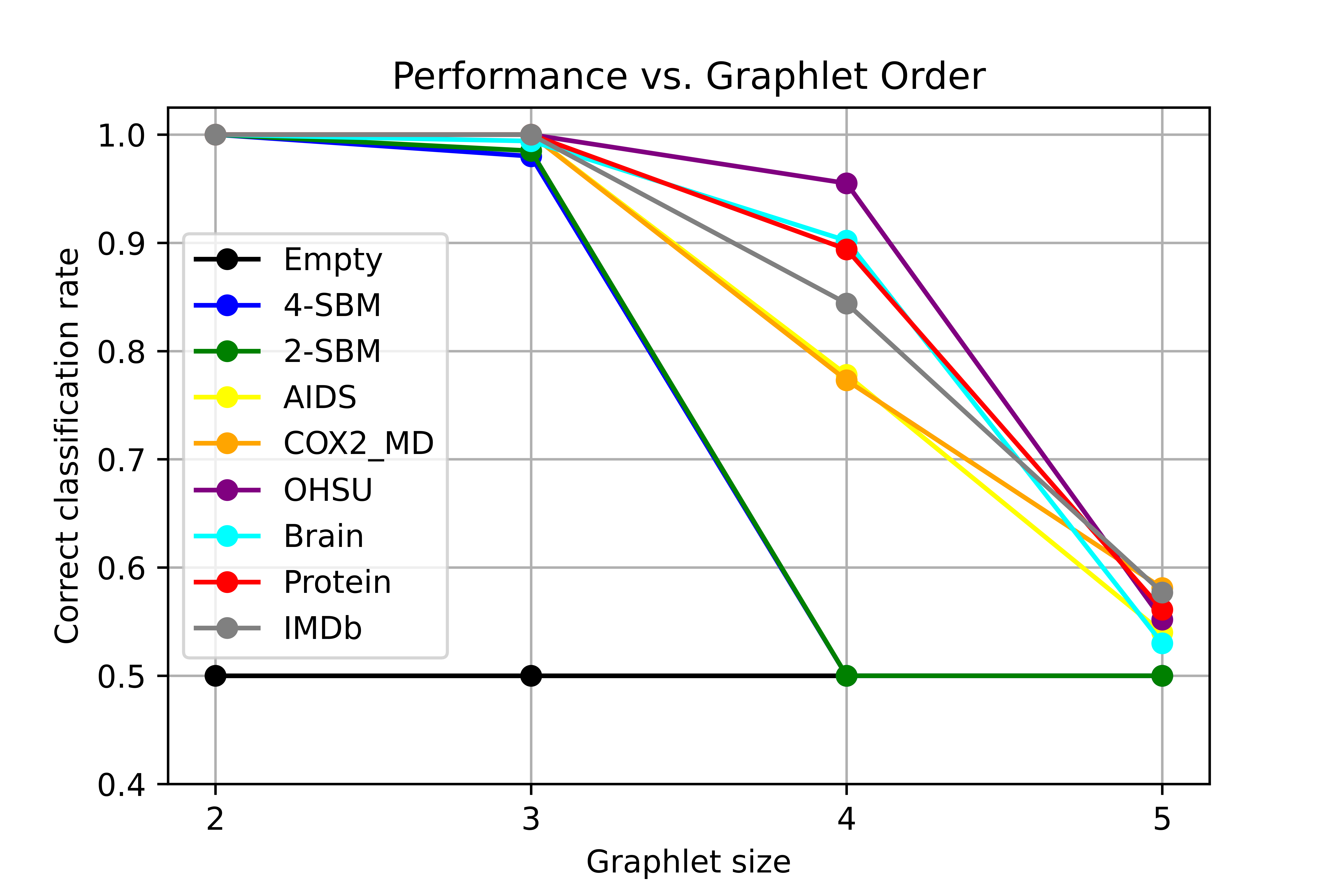}
\caption{Ability of the graphlet classifier to correctly separate real graphs from simulated graphs for each dataset. The given size of graphlets is the size used for training; the classifier used graphlets of one size larger.
}
\label{fig:order}
\end{figure}

After training, all classifiers performed poorly on the six real datasets. Results are summarized in Table \ref{tab:NN.classification}. Experiments generally used graphlets up to order 5, but results for smaller graphlets are depicted in Figure~\ref{fig:order}.

\begin{table}[h]
\centering
\begin{tabular}{l|cc}
Dataset & Graphlet & GNN \\ 
\hline
Empty graph & 0.500 & 0.500 \\
4-block SBM & 0.500 & 0.500 \\
2-block SBM & 0.500 & 0.500 \\
AIDS & 0.540 & 0.581 \\
COX2\_MD & 0.581 & 0.696 \\
OHSU & 0.552 & 0.579 \\
Brain & 0.530 & 0.539 \\
Protein & 0.561 & 0.573 \\
IMDb & 0.577 & 0.621 
\end{tabular}
\caption{
\label{tab:NN.classification}
This table lists the results of the two tests described in Section 4 for each dataset used. In each test, a classifier was trained to distinguish between real data and generated data. The given rate is how often the classifier correctly classified a graph, with an optimal value of 0.5.}
\end{table}

To further evaluate the model, we make use of the same Maximum-Mean Discrepancy (MMD) metrics introduced by \cite{you2018graphrnn}. The graph statistics used there are degree distributions, clustering coefficient, and orbit counts. A graph's clustering coefficient and orbit counts are equivalent to the ``triangle'' three-graphlet and the list of four-graphlets, respectively. On the two-block SBM and Protein datasets, we compare directly to GraphRNN \cite{you2018graphrnn}. Results can be found in Table 3 below.

\begin{table*}[h]
\begin{tabular}{c|ccc|ccc|ccc|ccc}
Dataset & \multicolumn{3}{c|}{four-block SBM} & \multicolumn{3}{c|}{AIDS} & \multicolumn{3}{c|}{COX2\_MD} & \multicolumn{3}{c|}{OHSU} \\ \hline
 & Deg. & Clust. & Orb. & Deg. & Clust. & Orb. & Deg. & Clust. & Orb. & Deg. & Clust. & Orb. \\
Self & 0.101 & 3.75e-3 & 6.84e-3 & 2.011 & 1.27e-2 &  & 0.401 & 1.17e-3 & 1.54e-3 & 0.352 & 2.67e-2 & 1.04e-2 \\
GraphMoE & 0.225 & 8.12e-3 & 2.06e-2 & 1.404 & 2.10e-4 & 1000 & 1.676 & 2.77e-2 & 5.01e-2 & 1.455 & 2.99e-2 & 8.83e-3 \\ \hline
Dataset & \multicolumn{3}{c|}{Brain} & \multicolumn{3}{c|}{IMDb} & \multicolumn{3}{c|}{two-block SBM} & \multicolumn{3}{c|}{Protein} \\ \hline
\multicolumn{1}{l|}{} & Deg. & Clust. & Orb. & Deg. & Clust. & Orb. & Deg. & Clust. & Orb. & Deg. & Clust. & Orb. \\
Self & 0.245 & 1.30e-2 & 5.39e-3 & 1.660 & 3.71e-2 & 1.45e-2 & 5.44e-2 & 4.10e-4 & 1.10e-3 & 4.20e-2 & 9.23e-3 & 1.45e-2 \\
GraphMoE & 0.251 & 5.44e-2 & 7.23e-3 & 5.681 & 8.69e-2 & 4.33e-2 & 0.475 & \textbf{9.37e-4} & \textbf{8.62e-3} & 0.767 & \textbf{3.88e-2} & \textbf{5.71e-2} \\
GraphRNN & - & - & - & - & - & - & \textbf{1.40e-2} & 2.00e-3 & 3.90e-2 & \textbf{3.40e-2} & 0.935 & 0.217
\end{tabular}
\caption{This table lists the values of three differentiation statistics for each real dataset. First, the dataset is split in half and the halves are compared. Then, the real data is compared to data generated by GraphMoE. For the 2-block SBM and Protein datasets, the statistic values for real data vs. data generated by GraphRNN are also given.}
\label{tab:results}
\end{table*}

\subsection{Results for extensions.}

In this section, we give additional results from applying various extensions and changes to the model. 

\subsubsection{\label{section:community}Community model.} 
We described an alternate version of the model tailored to graph data with community structure in section \ref{sec:community}. This version of the method learns only a small number of node representations, allowing it to be applied to datasets with significantly larger numbers of nodes, provided they exhibit the appropriate structure. Here, we  make use of the two-block SBM described in section \ref{sec:sbm}. We train the model using the RBF kernel with scale factor, up to graphlets of size five. The results are in Table~\ref{tab:community}.

The power of this alternative version of the model, however, is its ability to match graphs of much larger size. We increase the number of nodes to 10,000 and perform the same experiment. Since it is no more difficult to compute graphlets for small graphs than for large ones, the number of nodes included can be as large as desired. The results are again in Table~\ref{tab:community}.

\begin{figure}
    \centering
    \begin{subfigure}[b]{0.25\textwidth}
        \includegraphics[scale=0.2]{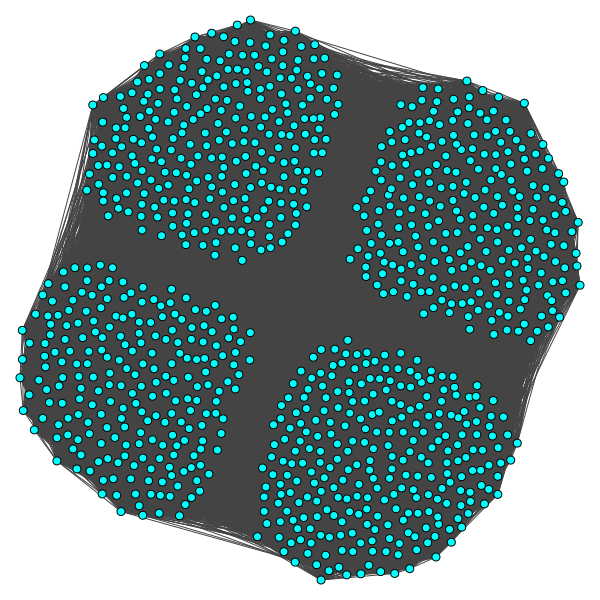}
        \caption{A graph produced by \\ the SBM.}
        \label{fig:gull}
    \end{subfigure}%
    \begin{subfigure}[b]{0.25\textwidth}
        \includegraphics[scale=0.2]{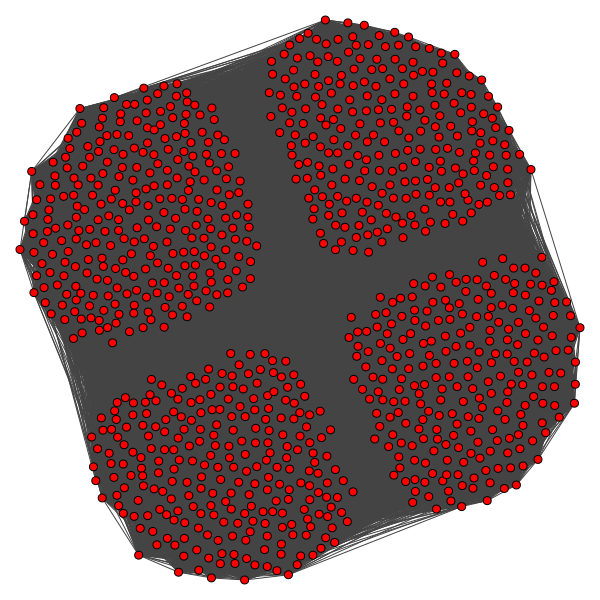}
        \caption{A graph produced by \\ the model.}
        \label{fig:graph_drawings_0}
    \end{subfigure}%
    \caption{\label{fig:compare}Comparison of graphs from the community model. The topological arrangement of nodes optimized by the \textit{igraph} package are quite similar. (Due to the very large number of edges, the connectivity structure cannot be observed at this image resolution.)
    }
\end{figure}

We also include visualizations of a larger graph generated from the SBM and from the model trained to imitate it; see Figure~\ref{fig:compare}. The drawings include 1000 nodes, as drawing more becomes computationally intensive. (\textit{Using} more is not any more difficult.) The visualizations were created using the \textit{igraph} package in Python.

\begin{table}[h]
\begin{tabular}{l|lll}
Dataset & Total Diff. & Max Diff. & Classification Rate \\ \hline
16 Nodes     & 0.0451           & 6.6e-3         & 0.557               \\
10,000 Nodes & 0.0194           & 9.4e-4         & 0.509        
\end{tabular}
\caption{Performance of community model trained on the four-block SBM. The classification rate is the rate at which the graphlet classifier could correctly separate SBM graphs from learned graphs, with optimal value 0.5.}
\label{tab:community}
\end{table}

\subsubsection{Inversely-weighted loss function.} 
For most datasets, there can be large variations among the frequencies of some graphlets of  given order. For instance, the AIDS dataset is quite sparse, and of all its subgraphs on 3 nodes, the subgraph containing one edge is ten times more present than the subgraph containing two edges. However, we find that these less common graphlets often contain much of the information about the overall random graph distribution, and matching them less than perfectly can have powerful effects on the results. 

In such cases, it may be preferable to reweight the loss function to increase the importance of rare graphlets. More precisely, we can replace the function $U(\theta) = |E_\theta (H) - \bar{H}|^2$ by
\[ U^*(\theta) = \sum_{F\in\mathcal{F}}\frac{(E_\theta (H_F) - \bar{H}_F)^2}{|\bar{H}_F|}. \]
This corresponds to using a diagonal matrix $D$ with diagonal elements given by $|\bar{H}_F|^{-1}$ in equation \eqref{eq:obj}.

The overall result is an increase in the ``Total Difference" between the target graphlets and the produced graphlets, but slight improvements in the ability of the model to fool the discriminators. Numerical results for three of the datasets are included in Table~\ref{tab:alt_loss} below.

\begin{table}[h]
\begin{tabular}{l|ll}
Dataset  & Total Difference & Classification Rate \\ \hline
AIDS     & 0.102            & 0.556               \\
COX2\_MD & 0.113            & 0.687               \\
Brain    & 0.157            & 0.515              
\end{tabular}
\caption{Results for the inversely-weighted loss function on three of the datasets, providing a slight improvement on the results in table \ref{tab:results}}
\label{tab:alt_loss}
\end{table}

\subsubsection{Use of larger graphlets.}

We now show how using partial graphlets (section \ref{sec:partial}) can help better fitting the degree distribution in our simulated graphs. (We note that 
GraphMoE is outperformed by GraphRNN in this metric on most datasets.) 

\begin{figure}
    \centering
    \begin{subfigure}[b]{0.95\textwidth}
        \includegraphics[scale=.6]{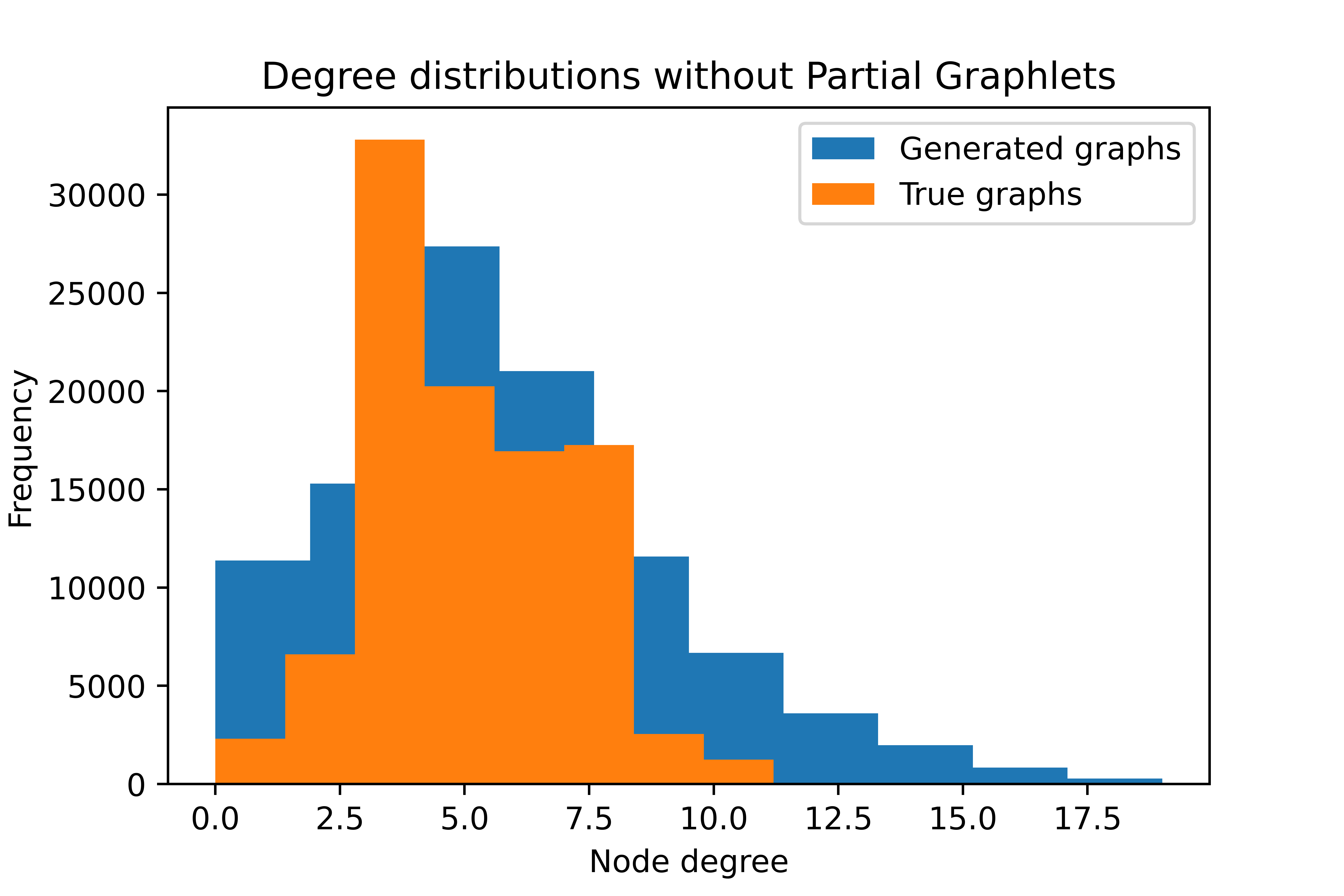}
    \end{subfigure}
    
    \begin{subfigure}[b]{0.95\textwidth}
        \includegraphics[scale=.6]{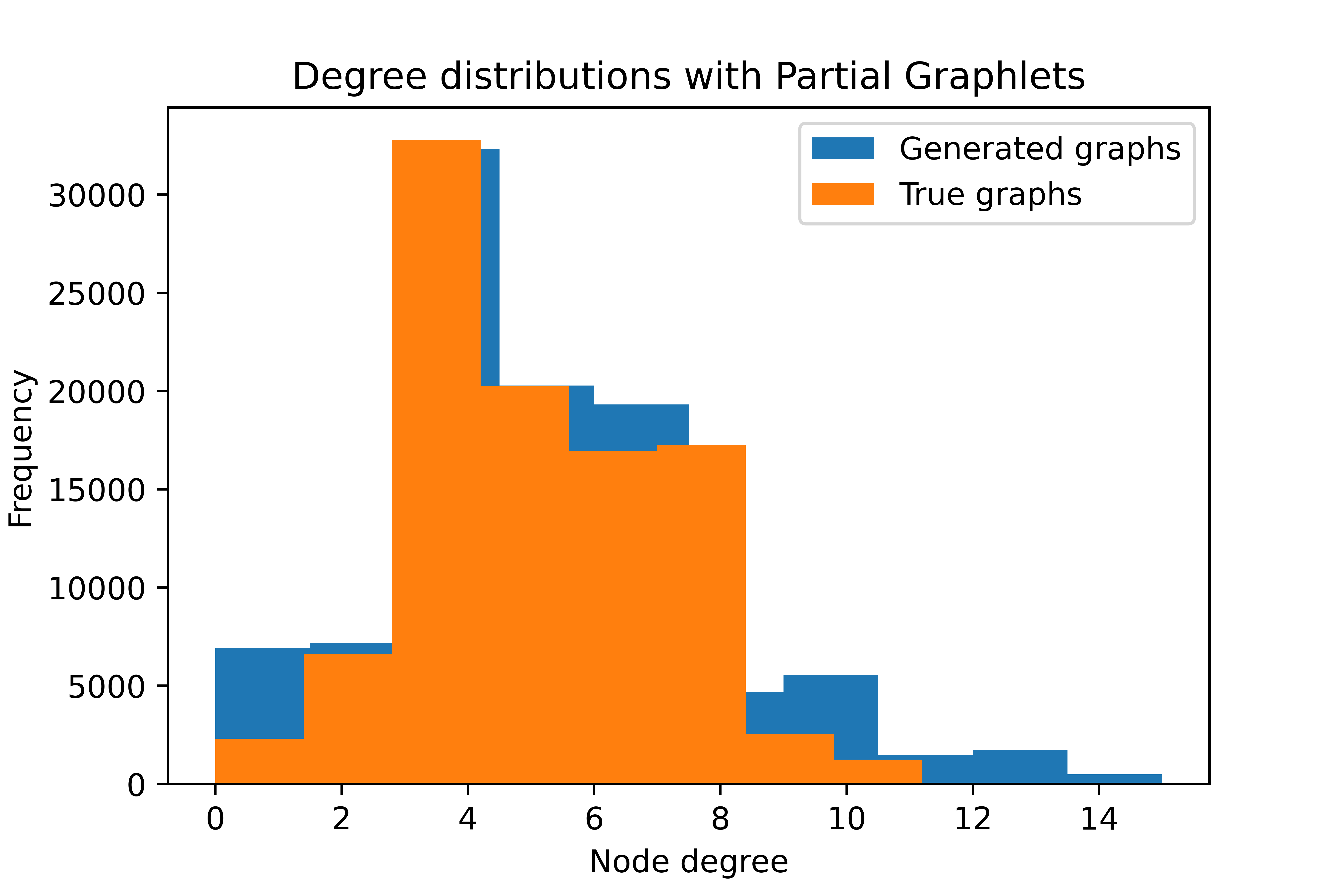}
        \label{fig:graph_drawings}
    \end{subfigure}%
    \caption{\label{fig:degrees}Histogram of degrees for nodes randomly sampled from a 4-block SBM (truth) and from GraphMoE trained with and without partial graphlets.
    }
\end{figure}

In Figure~\ref{fig:degrees} we display a comparison of the produced degree distributions to the ground truth for the two-block SBM and Protein datasets with and without use of the star partial graphlets for training. The addition of these higher-order graphlets does produce some visible improvement; however, it does not entirely fix the discrepancies. Going even further in the graphlet size would probably help, but  we are limited by computing power and training time.

\section{Discussion}
We have introduced, in this paper, a new  model of random kernel graphs and some variants, relying on a neural net generator. Our experimental results report good quality performance on several simulated and real datasets. Our model is able to imitate complicated classes of random graphs with a relatively simple architecture and small amount of computational time. The use of graphlets as the function $F$ ensures that the algorithm is able to properly capture distributions of graphs with unlabeled vertices.

There are two potential sources of variation in the graphs produced by the generator. The first is simply due to the nature of the RDPG; since each edge is generated randomly and independently, a variety of graphs can be produced using a single set of representations put out by the neural net. However, the use of $\omega$, a vector of Gaussian noise, as the input of the neural network creates further variation in the output of the network. Tests of variance were performed on the output of the neural network after being fed multiple inputs to determine to what extent this randomness is actually felt by the produced representations. We find that node embeddings tend to move around the embedding space greatly from run to run, with the variance of the embeddings' positions on average being around 30\% of their means. However, clusters of embeddings tend to move together, so that, depending on the kernel used, edge probabilities may remain more consistent despite the motion in response to $\omega$.

Computational costs for this method are quite low. While best performances are achieved after training times on the order of hours, fairly good performance begins to occur after just a few minutes of training. This allows for deeper or more complex neural network architectures to be employed if desired. 

Scalability is a desirable quality of generative models for random graphs; many datasets consist of large graphs, and the ability to add a great number of nodes is thus important. While our base model is most effective at capturing small-to-medium graphs, the community extension introduced in Section~\ref{sec:community} easily allows the model to include thousands of nodes.

The method's performance on disparate datasets is quite consistent. The graphs of the COX2\_MD dataset are nearly complete and node-homogeneous; in contrast, the Ego dataset is much sparser and has a central node. Despite these fundamental structural differences, the method's performance on each is similar (though the training speed for the COX dataset was much faster). The method appears robust enough to handle a variety of random graph distributions.

It remains unclear to what extent information contained in sets of graphlets is redundant. We see improved performance as the size of graphlets used to train is increased, and it seems obvious that the smallest graphlets alone do not contain enough information to completely specify a random graph distribution. The problem of reconstructing a single graph knowing only its graphlets of size $n-1$ is an open problem in graph theory, often called the Graph Reconstruction Conjecture \cite{reconstruction}. For our purposes, using graphlets of only size 5 and smaller is enough to fool discriminators (in particular, one that uses graphlets of size 6), so it is unclear how much information is added each time the size of graphlets is increased.

Future directions of research could include exploration of extensions designed to handle node- or edge-attributed graphs. Many real datasets, such as the visual scene graphs referenced in Section 1, include categorical attribute variables associated with nodes or edges. The relationships between edge probabilities and the attributes could again be captured using graphlets, so a variation of GraphMoE could readily be applied.

This method has potential to be applied in various fields of research where limited graph data is available. When a large amount of data is needed but acquiring it is cost-prohibitive, the method can be used to simulate additional data. Conditional distributions based on partially observed graphs can also be simulated empirically. Because of its low computational cost and high training speed, the method can easily be applied in computational vision, computational medicine, the study of social networks, or a number of other fields.

\printbibliography


\end{document}